\documentclass{article}

\usepackage{PRIMEarxiv}

\usepackage[utf8]{inputenc} 
\usepackage[T1]{fontenc}    
\usepackage{hyperref}       
\usepackage{url}  
\usepackage{float}
\usepackage{booktabs}       
\usepackage{amsfonts}       
\usepackage{nicefrac}       
\usepackage{microtype}      
\usepackage{lipsum}
\usepackage{amsmath}
\usepackage{fancyhdr}       
\usepackage{graphicx} 
\usepackage{algorithm}
\usepackage{algorithmic}
\usepackage{xcolor}
\graphicspath{{media/}}     
\usepackage{varioref}

\title{Evolutionary fine tuning of quantized convolution-based deep learning models
}

\author{
  Marcin Pietron \\
  AGH University\\
  Kraków, Poland\\
  \texttt{pietron@agh.edu.pl} \\
}

\begin{document}
\maketitle

\begin{abstract}
Deep learning models are the most efficient models in many machine learning tasks. The main disadvantage when using them in IoT, mobile devices, independent autonomous or real-time systems is their complexity and memory size. Therefore, much research has concentrated on compression techniques of deep learning architectures. One of the most popular technique is quantisation. In most of the works, the quantisation is done based on the nearest neighbour quantisation technique.
This work focuses on improving the quantisation efficiency in pretrained and quantised models. This approach has the potential to improve the final accuracy of quantised models. The main postulate of the work is that final quantisation states of the network based on nearest neighbour rounding does not guarantee optimal accuracy.
In the presented work, the neuroevolution strategy is used as an optimisation approach. The neuroevolution in each iteration changes the values of the small percentage of weights. It shifts theirs values to different quantisation states. The work shows that proposed neuroevolution with an appropriate set of operators and parameters can fast improve the accuracy of the quantised models. 
The results are presented for popular architectures such as VGG and Resnet for image classification and detection. Additionally, simulations were carried out for the autoencoder architecture.  
\end{abstract}

\keywords{deep learning \and quantization \and neuroevolution \and fine tuning}

\section{Introduction}
Low bit quantisation is a process which enables the adaption of deep neural architectures in embedded systems and helps to decrease theirs computational complexity. There are many approaches to decreasing bit precision while minimising the drop in accuracy \cite{motaz2020, han2015learning, kn:gysel, kn:anwar, pietron2019fast, kn:mishra2018, park2016faster, zhang2019, kn:park2017}. In \cite{pietron2023, motaz2020} authors show how quantisation can decrease execution time by reducing the bit-width format of deep learning models. Many of these techniques are based on complex optimisation algorithms which are time consuming and very often difficult for parallelisation. Creating a fast and efficient quantisation method which can improve the model accuracy presents a significant challenge. The quantisation methods can be divided according to several factors. The first factor is the stage in which it is performed. It can be done during the training process or just after it. The second factor is the type of quantisation. It can be linear or nonlinear quantisation. In this work, the presented method is run on linearly quantised model. 
The input of the presented algorithm is the pretrained and quantised model. The main assumption of the solution is that the quantization of the input model is done by rounding to the nearest neighbour. This is the approach used in most quantization algorithms \cite{motaz2020, han2015learning, gysel2016ristretto, kn:gysel, kn:park2017}.

In presented approach, the novel fine tuning approach is performed. It mutates the weights by value which is multiplication of least significant bit. The small percentage of weights are mutated in each iteration. The neuroevolution process is run on each layer separately. 
In addition, the algorithm is equipped with a sensitivity analysis of the tested model layers. It iterates through the layers from the least to the most sensitive. This feature allows to stop the process if some user defined drop in accuracy is achieved.
Another approach with a similar input assumption is the algorithm presented in the paper \cite{nagel2020}. It also elaborates that an approximation to the nearest neighbor does not guarantee optimal performance of the model. The main goal of the work was to create a methodology that is scalable and able to quickly improve accuracy of quantised models.

\section{Related work}

Quantisation is one of the most efficient techniques for the compression of deep learning models \cite{motaz2020,han2015learning}. Common strategies include the quantisation of all coefficients in a single layer with a specified number of bits to represent the integer and fractional components \cite{kn:anwar, kn:gysel} based on the range of values of the coefficients set. Another strategy is to represent coefficients and data by integer numbers with an appropriate scaling factor. Many  quantisation approaches in the literature adopt linear \cite{motaz2020, han2015learning, kn:gysel} or non-linear approaches including clustering \cite{pietron2019fast}. Quantisation can be performed during model training \cite{han2015learning} or can be run on a pre-trained model \cite{motaz2020, pietron2019fast}. Recently, several methods for low-bit representation have been designed \cite{kn:park2017, kn:zhang2018, kn:jung2018, kn:mcdonell2018, kn:mishra2018}. Many of these can not be run without some significant degradation in accuracy. One known advantage of quantisation is the fact that it facilitates the adoption of deep neural networks in specialised hardware accelerators with limited arithmetic bit-width and memory space \cite{xu2021, pietron2023}.

The authors in \cite{neuro_evolving} show that a two-level neuroevolution strategy scheme can outperform human-designed models in some specific tasks, for example, language modelling and image classification. 
In \cite{dimanov2021}, a novel neuroevolutionary method for the optimisation of the architecture and hyperparameters of convolutional autoencoders is presented. 
In \cite{okada2017}, it is shown that a genetic algorithm could evolve autoencoders that can reproduce the data better than manually created autoencoders with more hidden units. 
The first approach of co-evolutionary neuroevolution-based multivariate anomaly detection system is presented in \cite{faber2021}.  The authors show that proposed neuroevolutionary-based solution can outperform other human designed models on a well known benchmarks. These works show that neuroevolution can be very efficient method in deep learning architecture exploration.

Fine-tuning is quite a popular technique for improving the accuracy of the pretrained models. The most popular technique is gradient-based fine-tuning \cite{zhou2021}, \cite{ro2021}. The non-gradient approach is quite rare but can yield significant improvements as shown in \cite{nagel2020}. In \cite{nagel2020} the authors give the same input assumption is shown in this work. The authors in \cite{nagel2020} try to find the near optimal state of the model which allows to have quantisation weight values which are not rounded to the nearest neighbour.

In the presented work, a novel approach is proposed. The fine tuning based on neuroevolution is used for improving the accuracy of pretrained models.


\section{Quantisation}
After network distillation by the pruning process, quantisation can be performed as the next step of reducing the complexity of the model. Quantisation is the procedure of constraining values from a continuous set or a more dense domain to a relatively discrete set. It is possible to define general mapping from a floating-point data $x\in {\mathcal S}$ to a fixed-point $q\in\mathcal Q$ using a function $f_{\mathcal Q}: \mathcal S\rightarrow \mathcal Q$ as follows (assuming signed representation):
\begin{equation}
	\label{eq:quant}
	q = f_{\mathcal Q}(x) = \mu + \sigma \cdot \text{round}(\sigma^{-1}\cdot (x-\mu)). 
\end{equation}
In our case, $\mu=0$ and $\sigma = 2^{-{\bf frac\_bits}}$ where: 
\begin{equation}
	{\bf int\_bits}=\text{ceil}(\log_{2}(\max_{x\in{\mathcal S}} |x|)) 
\end{equation}

and 
\begin{equation}{\bf frac\_bits}={\bf total\_bits}-{\bf int\_bits}-1.
\end{equation}

The number of MAC operations in the layer is equal to $P_i \cdot C_i \cdot D_i \cdot H_i \cdot W_i$, where $P_i$ is the number of neurons in an output feature map, $C_i$ and $D_i$ are numbers of input and output channels, and $W_i$ and $H_i$ are the width and the height of the filter. The memory footprint of the single layer is $C_i \cdot D_i \cdot H_i \cdot W_i$. The quantised model has reduced the number of MAC operations. 
The same situation applies in the case of the memory footprint. The quantisation reduces the complexity further. The bit width of data decreases the number of cycles needed to run multiplication operations. In the case of the 8-bit quantisation (both weights and activations in 8-bit format), the number of MAC operations of the baseline floating point configuration is reduced by $1/9$. If weights are further reduced to 4-bit, the number of cycles in full 8-bit configuration are reduced by more than $1/2$. For 8-bit weights and 16-bit half precision activations, the reduction ratio is $2/9$. 

\section{Neuroevolution fine tuning}

Neuroevolution is an optimisation algorithm based on a genetic approach which helps to find more optimal neural architectures. In this work, the neuroevolution strategy is adapted for fine tuning the quantised weights in order to improve the accuracy of the model. The optimisation process is run on a pretrained quantised model. The pretrained model is quantised using linear quantisation with nearest neighbour rounding. The whole process in divided into two phases. The first phase is sensitivity analysis and the second is fine tuning the quantised layers by through the use of the neuroevolution approach.


\subsection{Sensitivity analysis}

The goal of the sensitivity analysis is to set up the ranking list of the most sensitive layers. The whole process is shown in Alg. \ref{alg:sensitivity}. 
At the beginning of the analysis, the empty list is created (line 1). The list will be used for storing the accuracy measured by quantising the specific layer. A single layer in each iteration  is then quantised into a low bit format (loop in lines 2-8). First, the copy of the original model is created (line 3). Then, the layer $l$ is quantised (line 4). The $\theta_l$ weight tensor is transformed to a low bit format $\theta_{l}^{q}$. The quantised layer is inserted to the whole model weight tensor (line 5). Finally, the inference is run in order to check the accuracy. The accuracy is added to the list (line 7). When all layers are run in the loop, the ranking is returned (line 9). 

\begin{algorithm}
\begin{algorithmic}[1]
\REQUIRE{$\psi$ -- desired bit-width}
\REQUIRE{$\Theta$ -- weights of a model}
\REQUIRE{$F_{\Theta}$ -- model}
\STATE{$\Lambda \gets \emptyset$} \COMMENT{List for top1 metric}

\FOR{$\theta_i$ $\textbf{in}$ $\Theta$}
\STATE{$F_{\Theta}' \gets copy(F_{\Theta})$}
\STATE{$\theta_{i}^{q} \gets$ q($\theta_i$, $\psi$=4)} \COMMENT{quantize layer to four bit}
\STATE{$\Theta' \gets \{\theta_0, \theta_1,...,\theta_{i}^{q} ,...,\theta_N\}$}
\STATE{$a_i \gets$ eval($F_{\Theta'}'$)}
\STATE{$\Lambda \gets \Lambda \cup a_i$}
\ENDFOR
\RETURN{argsort($\Lambda$)}
\caption{Sensitivity analysis}
\label{alg:sensitivity}
\end{algorithmic}
\end{algorithm}

The algorithm goes through all the layers and quantises them to a specified bit width - $\phi$. In the presented approach, the bit-width 
is set to 4. In line 1, the algorithm initialises the list in which it stores the accuracy values. 



\subsection{Fine tuning pretrained model}




The deep learning network is defined as a sequence of layers :
\begin{equation}
F_{\Theta}(X) = {f_{\theta_L}(f_{\theta_{L-1}}...(f_{\theta_{0}}(X)))}
\label{eq:model}
\end{equation}

The trainable parameters (weights) are defined as a following list:

\begin{equation}
\Theta = \{\theta_{0}, \theta_{1},...,\theta_{L}\}
\label{eq:theta_v}
\end{equation}

The quantized version of the weight tensor is as follows:

\begin{equation}
\Theta^q = \{\theta_{0}^q, \theta_{1}^q,...,\theta_{L}^q\}
\label{eq:Theta_}
\end{equation}

Each layer weight tensor can be quantized (using $q$ function) with specified bit width - $\phi$:

\begin{equation}
\theta_{i}^{q} = q(\theta_{i}, \phi)
\label{eq:theta_}
\end{equation}


The pretrained quantised model is represented as $F_{\Theta^{q}}$. It is an input to the neuroevolution strategy which is described in Alg.\ref{alg:neuroevolution}. The algorithm iterates through all the layers in the model (line 1-20). It begins from the less sensitive layers. It takes the weight tensor of the layer (line 2) and adds the copies of it to the initial population $P$ (line 4). The initial population is just a list of copies of the quantised $l$ layer:

\begin{equation}
P = \{\Theta^{q}_{l},\Theta^{q}_{l},...,\Theta^{q}_{l}\}
\label{eq:pop}
\end{equation}

Then, mask is generated using binomial distribution (line 9). The mask is a binary tensor:

\begin{equation}
M_i \in \{0,1\}^{SH_{\theta_{i}}}
\label{eq:m}
\end{equation}

The shape $SH_{\theta_i}$ of the mask is the same as the shape of weight tensor $\theta_i$. The mask has $p$ percentage of $1$ values. The weights position for which the mask is set to $1$ will be change during the mutation step (line 12): 


\begin{equation}
\theta_{i} = \theta_{i} + M_i \odot \vec{q}
\label{eq:theta}
\end{equation}


The vector $\vec{q}$ is generated by following equation (line 11): 

\begin{equation}
\vec{q} = \vec{r} \cdot \sigma_l
\label{eq:q}
\end{equation}

where $sigma_l$ is least significant bit value for the specific $l$ layer and $\vec{r}$ is defined as:

\begin{equation}
\vec{r} = \{r_0, r_1,...,r_N\}
\label{eq:r}
\end{equation}

Each $r_i$ value is a randomly generated number from the predefined set (line 10): 

\begin{equation}
r_i \in \{-2, -1, 1, 2\}
\label{eq:el}
\end{equation}

The $-1$ and $1$ values are generated with probability equal to 40\% each. The $-2$ and $2$ are generated with 10\% each. Finally the mutated layer is added to the model weight tensor (line 15) and is evaluated. In line 18, the accuracy values in the population are sorted. The best candidates are taken to the next iteration (line 19). 


\begin{algorithm}
\begin{algorithmic}[1]
\REQUIRE{$r$ -- ranking list}
\REQUIRE{$P_S$ -- population size}
\REQUIRE{$\Theta$ -- weights of a model}

\FOR{$l$ $\textbf{in}$ $r$}
\STATE{$P \gets \emptyset$}
\FOR{$i$ $\textbf{in}$ $P_S$}
\STATE{$P \gets P \cup copy(\theta^{q}_{l})$}
\ENDFOR
\STATE{$A \gets \emptyset$}
\FOR{$i$ $\textbf{in}$ $I$}
\FOR{$j$ $\textbf{in}$ $P_S$}
\STATE{$M_l \gets \{B(n,p)\}, B(n,p)\},...,B(n,p)\}$} \COMMENT{Bernoulli distribution}
\STATE{$\vec{r} \gets$ sample vector from $\{-2,-1,1,2\}$} 
\STATE{$q = \vec{r} \cdot \sigma_l$}
\STATE{$P[j] = P[j] + M_l \odot q$}
\STATE{$P \gets P \cup P[j]$} \COMMENT{add mutated layer to the population}
\STATE{$\theta_l=P[j]$}
\STATE{$\Theta = \{\theta_0,...,\theta_l,...,\theta_N\}$}
\STATE{$A \gets A \cup eval(F_{\Theta})$}
\ENDFOR

\STATE{$sorted \gets argsort(A)$}
\STATE{$P \gets P[sorted[:P_S]]$} \COMMENT{take $P_S$ best candidates}
\ENDFOR
\ENDFOR
\RETURN{$P$}
\caption{Fine tuning}
\label{alg:neuroevolution}
\end{algorithmic}
\end{algorithm}





\section{Results}

The presented results show that the proposed solution can significantly improve the results achieved by quantisation based on rounding to the nearest neighbour. The baseline floating point results are shown in Table \ref{baseline}. There are four models and four datasets. The FasterRCNN is a object detection model. It is tested on a VOC Pascal dataset. The baseline mAP (mean average precision) is 71.2\%. The remaining three models: Resnet18, VGG16 are the models for image classification and CNN autoencoder - 12 layer autoencoder from \cite{faber2021} for time series anomaly detection. The Resnet18 and VGG16 are simulated on CIFAR100 and ImageNet. The CNN autoencoder is tested on SWAT benchmark. For CIFAR100 and ImageNet the metric is top1. The metric for SWAT is F1.



\begin{table}[h]
\centering
\caption{Baseline results}
\begin{tabular}{|c|c|c|c|c|c|c|}
\hline
\textbf{} & \textbf{CIFAR100} & \textbf{VOC Pascal} & \textbf{ImageNet}  &  \textbf{SWAT} \\ \hline
\textbf{FasterRCNN}     &    -      &         71.18             &   - & \\ \hline
\textbf{Resnet18}     &     75.0      &          -            &   69.75 & \\ \hline
\textbf{VGG16}     &       70.4       &       -    &  71.59  & -\\ \hline
\textbf{CNN AE}     &      -      &    -       &    -  &  82.0 \\ \hline

\end{tabular}
\label{baseline}
\end{table}

\begin{table}[h]
\centering
\caption{8 bit linear quantisation results}
\begin{tabular}{|c|c|c|c|c|}
\hline
\textbf{} & \textbf{CIFAR100} & \textbf{VOC Pascal} & \textbf{ImageNet} & \textbf{SWAT}\\ \hline
\textbf{FasterRCNN}     &      -        &              71.04        & - & -\\ \hline
\textbf{Resnet18}     &     74.9       &   -                &  69.5 & -\\ \hline
\textbf{VGG16}     &     70.4       &    -       &    71.25 & -\\ \hline
\textbf{CNN AE}     &    -       &    -       &    -  & 81.98\\ \hline

\end{tabular}
\label{eight_bit}
\end{table}

\begin{table}[h]
\centering
\caption{Linear quantisation results: 4 bit Resnet18 and VGG16 for CIFAR100, 6 bit Resnet18 and VGG16 for ImageNet, 5 bit CNN AE, 6 bit FasterRCNN}
\begin{tabular}{|c|c|c|c|c|c|}
\hline
\textbf{} & \textbf{CIFAR100} & \textbf{VOC Pascal} & \textbf{ImageNet} & \textbf{SWAT} \\ \hline
\textbf{FasterRCNN}     &       -      &                 64.0     &    - & - \\ \hline
\textbf{Resnet18}     &      73.5      &     -                 &   62.04 & - \\ \hline
\textbf{VGG16}     &      68.9    &      -     &   54.73 & -\\ \hline
\textbf{CNN AE}     &     -       &    -       &    - &  80.70 \\ \hline

\end{tabular}
\label{four_bit}
\end{table}

In Table \ref{eight_bit} 8-bit quantisation results for all models are presented, in Table \ref{four_bit} 4-bit Resnet18 and VGG16 on CIFAR100, 6-bit Resnet18 and 5-bit VGG16 on ImageNet, 6-bit FasterRCNN and 5-bit CNN AE results are described. In Table \ref{eight_bit} and \ref{four_bit} linear quantisation with rounding to nearest neighbour was used (section 3). It can be observed that in the case of 8 bit quantisation, a less than 1\% drop in accuracy was achieved in all models and datasets. 
In the case of lower bit quantisation, the drop is significantly higher for 6-bit Resnet18 (7.71\%) and 5-bit VGG16 (16.86\%) on ImageNet. The same can be observed for 6-bit FasterRCNN where the drop is 7.18\%. The small drop is for 5-bit CNN AE (1.3\%), and for Resnet18 and VGG16 on CIFAR100 (less than 1.5\%). In Table \ref{eight_bit_ft} and \ref{four_bit_ft} the results are presented after applying fine tuning neuroevolution approach. In simulations for Resnet18 and VGG16, the population size was set to 32, the number of iterations was 64, and 2\% of weights were mutated in each candidate. For CNN AE the population was 24, iterations number was set to 30. In the case of FasterRCNN the simulations were run with 16 iterations and 16 solutions in population. The FasterRCNN achieves a drop in accuracy of less than 1\% in the cases of both 6 and 8-bit quantisation. The fine tuned 8 bit Resnet18 and VGG16 have better results on CIFAR100 than floating point version (+0.2\%). The 4 bit fine tuned versions are just 0.2\% worse than baseline floating point counterparts. For CNN autoencoder the accuracy is very close the floating point version (82.09\% for 8 bit, which is better by 0.09\%, and 81.93\% for 5 bit version, which is better by more than 1\% from the baseline 5 bit quantisation).  
The highest drops in accuracy are observed for 6-bit and 5-bit quantisation for ImageNet, but the proposed fine tuning significantly improves the baseline quantisation results (by more than 6\% in case of Resnet18 and more than 14.5\% in case of VGG16) and achieves accuracy about 1.5\%  and 2\% worse than floating point baseline versions for Resnet18 and VGG16, respectively. For 8-bit quantisation the results are also improved and very close to the floating point results (less than 0.1\%).

The obtained results show that the proposed method allows to achieve results very close to the original non-quantized model. Only for the ImageNet dataset it gives worse results than the method presented in \cite{nagel2020}, which is able in some cases to regain the accuracy in 4 bit format. The main advantage of proposed method is definitely better scalability and better time efficiency than \cite{nagel2020}.

\begin{table}[ht]
\centering
\caption{8 bit fine tuned results}
\begin{tabular}{|c|c|c|c|c|c|}
\hline
\textbf{} & \textbf{CIFAR100} & \textbf{VOC Pascal} & \textbf{ImageNet} & \textbf{SWAT}\\ \hline
\textbf{FasterRCNN}     &       -       &              71.34        & - & -\\ \hline
\textbf{Resnet18}     &      75.2      &    -                  &   69.7 & - \\ \hline
\textbf{VGG16}     &     70.6      &     -      &    71.5 & - \\ \hline
\textbf{CNN AE}     &    -       &    -       &    - &  82.09 \\ \hline

\end{tabular}
\label{eight_bit_ft}
\end{table}

\begin{table}[ht]
\centering
\caption{Fine tuned linear quantisation results: 4 bit Resnet18 and VGG16 for CIFAR100, 6 bit Resnet18 and VGG16 for ImageNet, 5 bit CNN AE, 6 bit FasterRCNN}
\begin{tabular}{|c|c|c|c|c|c|}
\hline
\textbf{} & \textbf{CIFAR100} & \textbf{VOC Pascal} & \textbf{ImageNet} & \textbf{SWAT} \\ \hline
\textbf{FasterRCNN}     &     -      &               70.5       &  - & -\\ \hline
\textbf{Resnet18}     &    74.8     &     -                 &   68.2 & - \\ \hline
\textbf{VGG16}     &     70.2     &      -     &   69.6  & -\\ \hline
\textbf{CNN AE}     &   -     &      -     &   -  &  81.93 \\ \hline

\end{tabular}
\label{four_bit_ft}
\end{table}


\newpage

\section{Conclusions and future work}
The results described in this work show that the proposed solution is quite efficient. The simulations run on different datasets with various models showing that evolutionary fine tuning can boost accuracy and minimize the drop after nearest neighbour quantisation. This is a good alternative for gradient fine tuning. The next advantage is its scalability. It can be fully parallelised. In the next step the algorithm will be incorporated with the new features like finding different bit formats for each channel in a single layer. Future works will concentrate on improving the method by combining gradient with non-gradient fine tuning and adapting similar methodology for nonlinear quantisation. The more architectures like MobileNet and Transformer-based will be explored.

\bibliographystyle{unsrt}  
\bibliography{references}

\end{document}